\documentclass[format=acmsmall, review=false, screen=true]{acmart}

\renewcommand\footnotetextcopyrightpermission[1]{} % removes footnote with conference information in first column
\usepackage{booktabs} % For formal tables
\usepackage{graphics} % for pdf, bitmapped graphics files
\usepackage{amsmath} % assumes amsmath package installed
\usepackage{amssymb}  % assumes amsmath package installed
\usepackage{amsfonts}
\newtheorem{problem}{Problem}

\usepackage{algorithm}
\usepackage{algpseudocode}
\usepackage{subfigure}
\usepackage{etoolbox}

\begin{document}
\title{Gray-box Adversarial Testing for Control Systems with Machine Learning Component\footnote{}}
\thanks{*A version of this paper will be presented in International Conference on Hybrid Systems: Computation and Control (HSSC) 2019, Montreal, Canada.}
\author{Shakiba Yaghoubi}

\affiliation{%
\institution{SCIDSE, Arizona State University}
\city{Tempe}
\state{Arizona}
\postcode{85281}
}
\email{syaghoub@asu.edu}

\author{Georgios Fainekos}
\affiliation{%
\institution{SCIDSE, Arizona State University}
\city{Tempe}
\state{Arizona}
\postcode{85281}}
\email{fainekos@asu.edu}

\begin{abstract}
Neural Networks (NN) have been proposed in the past as an effective means for both modeling and control of systems with very complex dynamics. 
However, despite the extensive research, NN-based controllers have not been adopted by the industry for safety critical systems.
The primary reason is that systems with learning based controllers are notoriously hard to test and verify. 
Even harder is the analysis of such systems against system-level specifications. 
In this paper, we provide a gradient based method for searching the input space of a closed-loop control system in order to find adversarial samples against some system-level requirements.
Our experimental results show that combined with randomized search, our method outperforms Simulated Annealing optimization.
\end{abstract}




\maketitle
\section{Introduction}

There is a long history of investigating the application of Neural Networks (NN) in high assurance systems \cite{SchumannL2010book}.
The advantages of including a NN in the control loop can be substantial.
For example, a system may include components with complex dynamics that cannot be modeled by first principles and need to be learned.
Most importantly, a high assurance system needs to be able to adapt in catastrophic situations. 
NNs provide such an adaptation mechanism with only limited assumptions on the structure of what is to be learned.
However, even though there has been substantial progress in the stability analysis and verification of such systems \cite{NguyenJ2010chapter}, the problem of system level verification of transient system behaviors still remains a major challenge in these systems.

In this paper, we report on progress on the automatic generation of adversarial test cases (falsification) for nonlinear control systems with NN components in the loop. 
%Real time properties of systems can be specified using different logics \cite{chapter5,mehrabian2017timestamp}. 
We assume that the properties of systems -that can be specified using different logics (e.g, \cite{chapter5,mehrabian2017timestamp})- are expressed in Signal Temporal Logic (STL) \cite{chapter5}, and we develop a framework that searches for adversarial tests through functional gradient descent. 
In particular, we propose using a local optimal control based search combined with a global optimizer since the resulting optimization problem is non-convex. 

We remark that our approach neither requires analytical information about the system model nor the NN architecture.
However, our framework requires and utilizes information readily available by most model based development tools for control systems.
Namely, it requires linearizations of the closed loop system at given operating points.
The linearizations help us approximate the gradient descent directions without the need for computing sensitivity matrices or numerical approximations of the descent directions.

While our approach does not generalize to falsification of systems with any kind of neural network, e.g., NN with non-smooth activation functions, it can be used for systems that contain Recurrent Neural Networks (RNN) with smooth activation functions. 
Typically, RNN cannot be handled by the existing testing and verification methods.
Finally, we remark that our proposed method could be extended to hybrid control systems with NN with Rectified Linear Units (ReLU) if results similar to \cite{YaghoubiF2017cyphy,zutshiDSK2014emsoft} are utilized.  

{\bf Summary of contributions:}
We develop an adversarial test generation (falsification) framework for control systems with RNN in the loop based on optimal control theory. Unlike our previous works \cite{yaghoubi2018falsification,yaghoubi2017hybrid} in which the input signal is parameterized using finite number of parameters, in this work the input is calculated using an optimal-control approach which searches directly in the infinite search space of the input functions.
We experimentally demonstrate that our framework vastly outperforms black-box system testing methods.
Namely, in our case study, the proposed framework always returns falsifications when the black-box methods fail to do so.

%\todo[inline]{@Shakiba: please verify and expand the above if necessary}

% Similar to the work in \cite{lioptimal} we use the well known results in optimal control theory but for systems equipped with neural networks and against properties that are not necessarily described in the form of standard optimal control cost functionals.  

\section{RELATED WORK}
% One of erkan's work and say it's stochastic search
% Derrosi's work but they are static and see erkan's discussion
% Mine that use approximation
%
The importance of formally analyzing safety-critical systems involving machine learning components is discussed in \cite{seshia2016towards}. It's been shown in \cite{szegedy2013intriguing} that even well-trained NNs can be fooled by applying adversarial perturbations to their inputs. While many other works like \cite{carlini2017towards,huang2017safety} have studied different properties of Machine Learning components in isolation, only recently these components are studied in the context of closed loop dynamical systems (e.g, \cite{dreossi2017compositional}). 
For example, in terms of testing and verification for NN, in \cite{huang2017safety}, a verification framework based on Satisfiability Modulo Theory (SMT) for feed-forward multi-layer NNs is developed. The framework aims to evaluate the robustness of image classifiers to manipulations. The work is limited to feed-forward networks and proves properties statically, in other words, it does not consider the NN in a closed loop system.

We highlight that our work utilizes ideas from falsification methods based on optimal control \cite{lioptimal,yaghoubi2017hybrid,AbbasWFJ14acc,zutshiDSK2014emsoft}. 
Similar to the work in \cite{lioptimal}, we use well known results in optimal control theory but for systems equipped with NN and against properties that are not necessarily described in the form of standard optimal control cost functionals.
In \cite{AbbasWFJ14acc}, authors study the falsification problem of white-box nonlinear systems with piece-wise constant inputs. They calculate gradient descent directions for system's initial conditions and input parameters using sensitivity analysis. This work has been extended in \cite{yaghoubi2017hybrid} to include gray-box nonlinear systems using the linearizations of the system model, and in \cite{YaghoubiF2017cyphy} to include hybrid systems. Another search-based method for falsification of Hybrid systems is multiple shooting optimization technique which is studied in
\cite{zutshiDSK2014emsoft}.

In \cite{dutta2017output}, a tool for calculating the reachable set of Rectified Linear Units (ReLU) feed-forward NNs based on mixed integer linear programs is introduced. This tool is used in \cite{dutta2018learning} for verifying stability and reachability properties of systems with feed-forward NNs in a feedback loop.
%While most of the recent work concentrates on neural networks with RelU activation functions, based on the well-known survey \cite{hunt1992neural},
However, NNs in complex dynamical systems may contain more complicated activation functions like sigmoids, tanh or Gaussian functions other than ReLU. 

%\todo[inline]{@Shakiba: I would replace the "usually" above with "may contain more". These days ReLU is the bread and butter}

Along these lines, the authors in \cite{xiang2018reachability} study the problem of safety verification for dynamical systems with feed-forward NNs with general activation functions. They develop an algorithm to compute an over-approximation of the reachable set of the neural network control
system over a finite time horizon.

Due to the complexity and the growth in the model order, formal verification methods cannot currently be used for control systems with more general NNs than feed-forward. In these applications one need to resort to testing methods. 
The works in \cite{abbas2017safe,tuncali2018simulation,DreossiEtAl2017rmlw, dreossi2017compositional} test autonomous vehicles equipped with NNs for perception, guided by system-level requirements.

\section{PRELIMINARIES}
\subsection{Neural Networks}
Neural Networks (NN) are brain-inspired functions/dynamical systems that can learn to replicate real systems if provided by enough data about that system. NN's consist of input, output and usually hidden layers that each include a number of nodes/neurons connected to transform the input into a suitable signal for replicating the desired output. The input layer passes the inputs to the network, where some computations are applied on them in the hidden layers, and the output layer consists of at least one node that generates the output vector. 
The inputs to each node are the outputs from other nodes, and the output of each node is computed by applying nonlinear functions to the weighted sum of its inputs.
Many methods have been studied in literature to train a NN  to replicate a system's behavior, most of which minimize a loss function, such as the mean-squared error of the output. We briefly introduce two types of the most generally used NNs in the following: 
\subsubsection{Feed forward Neural Networks (FNN)}
FNNs are the simplest type of NNs. They are static or memory-less networks with no feedback loops. Multi-layer perceptron (MLP) is the most general form of FNN, which has the ability to approximate any nonlinear function (Universal Approximation Theorem \cite{hornik1989multilayer}).

% Assuming $l$ layers in the FNN, the jth node in the ith layer with $n_i$ total number of neurons, applies the following function to its inputs,

% \begin{equation}
%     y_{ij} = \phi_i(W_{ij}^T u_i+b_{ij}) \;\;\; j\in[1,n_i], i\in[1,l]
% \end{equation}
% where assuming that there are $m_i$ inputs to the neuron, $u_i\in R^m_i$ represents the inputs vector, $W_{ij}\in R^m_i$ is a weight vector, $b_{ij} \in R $ is a bias, $y_{ij}\in R$ is the output of the neuron, and $\phi_i: R\rightarrow R$ is an activation function which usually is one the continuous nonlinear functions: ReLU, tanh, arctan, logistic, Softmax, etc. 
Assuming $l$ layers in the FNN, the ith layer applies the following function to its inputs $u_i\in {\rm I\!R}^{m_i}$,

\begin{equation}\label{fnn}
    y_{i} = \phi_i(W_{i}^T u_i+b_{i}) \;\;\; i\in\{1,2...,l\}
\end{equation}
where assuming that the layer has $n_i$ outputs $y_{i}\in {\rm I\!R}^{n_i}$ (usually $n_i = m_{i+1})$, $W_{i}$ is a ${\rm I\!R}^{m_i}\times {\rm I\!R}^{n_i}$ weight matrix, $b_{i} \in {\rm I\!R}^{n_i} $ is a bias vector, and $\phi_i: {\rm I\!R}^{m_i}\rightarrow {\rm I\!R}^{n_i}$ is an activation function which is usually one of the continuous nonlinear functions: ReLU, tanh, arctan, logistic or sigmoid.
%$\it{softmax}(y) = \frac{\Sigma_{i = 1}^n y_ie^{\alpha y_i}}{\Sigma_{i = 1}^n e^{\alpha y_i}}$ which approaches to the max function as $\alpha \rightarrow \infty$.
  The weight matrices $W_{i}$ and the bias vectors $b_i$ should be adjusted using a training approach \cite{haykin2009neural}. After the training phase, the function  $\it{FNN}:{\rm I\!R}^{m_1}\rightarrow {\rm I\!R}^{n_l}$ formed by neurons of Eq. (\ref{fnn}), calculates the final output of the feed forward neural net at time t given the input at that time: $y(t) = \it{FNN}(u(t))$

% Despite the impressive ability of approximating nonlinear functions, much complexities represent in predicting the output behaviors of MLP (5
\subsubsection{Recurrent Neural Networks (RNN)}\label{RNNexp}
Unlike FNNs, RNNs are dynamic networks. The feedback loops between neurons equip the network with long/short term memory. The output at each time $t$ represented as $y(t) = \mbox{RNN}(t,u(.))$ is a function of the vectorized input signal/sequence $u(.)$\footnote{Because of the properties of RNN, the output at each time $t'$ can be dependent on the values of the input signal at any time $t<t'$, so instead of using $u(t)$ that represents the value of signal at time t, we use $u(.)$ to represent the input trajectory} and is a solution to the following continuous or discrete system of equations:
\begin{align}
\label{RNN}
    &\dot{x}_{nn}=f^r_c(x_{nn},u), \;\; \mbox{or}\nonumber\\ x_{nn}(t)=f^r_d&(x_{nn}(t-1),x_{nn}(t-2),...,u(t))\\
    &y(t) = g(x_{nn}(t)) \nonumber
\end{align}
where $x_{nn}$ is the internal state (memory) of the {\it RNN} which is usually initially zero ($x_n(0) = 0$). These states are the outputs of the delay/integrator blocks whose inputs are  calculated using the functions $f_c^r$ or $f_d^r$ given the input and (previous) states.
Note that the above formulation describes the overall input output relationship of the {\it RNN} rather than the individual neurons. The {\it RNN} output at each time $t$ is a function of the states $x_{nn}$ at $t$.

We denote the solution of an arbitrary NN at time $t$ as $\it{NN}(t,u(.))$.
\subsection{Closed-loop Control Systems Description}
\begin{figure}
\centering
\includegraphics[width=.63\linewidth]{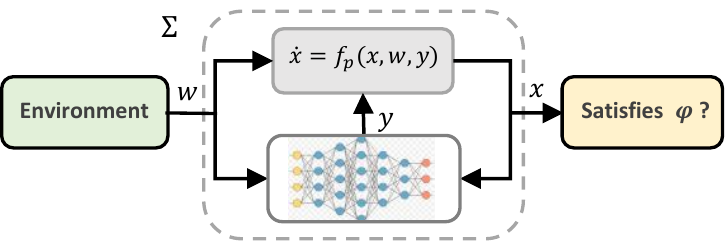}
\vspace{-5pt}
\caption{A close-loop control system containing a NN}
\vspace{-5pt}
\label{fig:1}
\end{figure}

In this paper NNs can be combined with a system plant in a general way. Many of the dynamical systems in which NNs are used for controls (in feedback, feedforward or end-to-end), unmodeled dynamics estimation or predictions, can lie under the class of systems that we consider (shown in Fig. \ref{fig:1}). The system is studied in the bounded time interval $[0,T]$ and described in the following.
\begin{align}\label{eq1}
    \Sigma:\; \dot{x}_p&=f_p(x_p,w,\mbox{NN}(t, x_p(.) ,w(.)))
\end{align}
where $x_p\in X \subset {\rm I\!R}^n$, $x_p(0)\in X_0$, and $w\in U \subset {\rm I\!R}^m$ are the system states, state initial values, and inputs, respectively. Also, $x(.), w(.)$ are the state and input trajectories, $\mbox{NN}:{\rm I\!R}_+\times X^{[0,T]}\times U^{[0,T]}\rightarrow {\rm I\!R}^k$, and $f:{\rm I\!R}^n\times {\rm I\!R}^m\times {\rm I\!R}^k \rightarrow {\rm I\!R}^n$ is a continuous function. 
The solution to system (\ref{eq1}) at time $t$ with initial condition $x_p(0)$ is denoted by $s(t,x_p(0),w)$.

\subsection{Specifications}
Desired system behaviors can be specified using  Signal Temporal logic (STL) formulas which are reviewed in \cite{chapter5}. These formulas are created by combining {\it{atomic propositions}} or {\it predicates} using logical and temporal operators. Logical operators include: {\it{and}} ($\wedge$), {\it{or}} ($\vee$), and {\it{not}} ($\neg$), and  temporal operators include: {\it{always}} ($\Box$),{ \it{eventually}} ($\Diamond$), and {\it until} ($\mathcal{U}$) that can be combined with time intervals to specify when operators are active.

Given the system state trajectory $s(t,x_0,w)$, a robustness value can be calculated with respect to an STL formula $\varphi$ (see \cite{FainekosP09tcs}), which shows how well the trajectory satisfies the formula. Positive values indicate satisfaction and negative values indicate violation. The absolute value of the robustness shows how far the trajectory is from being satisfied/falsified. 

The robustness value is calculated using max and min functions over the distances of the points on the trajectory from sets that are defined by the formula predicates and as a result the robustness function is not differentiable. In \cite{pant2017smooth} differentiable semantics of logic are defined approximately. The accuracy of the approximation however depends on various parameters and there is not a mature enough tool to calculate the robustness using them yet either. So in the following, we use the approach in \cite{AbbasWFJ14acc} to deal with the non-differentiability of the robustness function:

It can be shown that the absolute value of the robustness of the trajectory $s(t,x_p(0),w)$ corresponds to the distance between a point $s(t_*, x_p(0), w)$ on the trajectory and a point $z_*$ that belongs to a critical set. The critical set corresponds to a predicate in the STL formula $\varphi$, and $t_*$ is called the critical time. The variables $z_*$ and $t_*$ are simply calculated using tools such as S-Taliro while evaluating the robustness. The robustness of neighboring trajectories $s(t,x'_p(0), w')$ where $x'_p(0) = x_p(0)+\delta x_p(0)$, and $ w'(t) = w(t)+\delta w(t)$ is upper bounded by $||s(t_*,x'_p(0),w')-z_*||$, so minimizing Eq. \ref{e2} w.r.t. $x'_p(0)$ and $w'$ will locally minimize the robustness function. Note that cost function depends on $x_p(0)$ and $w$ through $z_*$ and $t_*$.
\begin{equation}
\label{e2}
J_{x_p(0),w} = \frac{1}{2}\big( s(t_*,x'_p(0),w')-z_*\big) ^\top\big(s(t_*,x'_p(0),w')-z_*\big)
\end{equation}

\section{ANALYTICAL ADVERSARIAL TESTING}
\subsection{Problem Formulation}
In adversarial testing, we are interested in finding adversarial $w\in U^{[0,T]}$ and $x_0\in X_0$ for which the solution to the system (\ref{eq1}) does not satisfy a given formula $\varphi$. The adversary can be used later to improve the system performance by adapting or retraining the NN (Similar to \cite{dreossi2018counterexample}). We look at the problem as a constrained optimization problem in which we minimize the robustness function over $X_0$ and $U^{[0,T]}$ and under the dynamics of Eq. (\ref{eq1}). This optimization problem can be locally solved by minimizing the cost in Eq. (\ref{e2}) instead of the robustness value. Also, we integrate the NN with the plant and rewrite the system in Eq. (\ref{eq1}) as:
\begin{equation}
\label{cl}
    \dot{x} = f(x,w)
\end{equation}

Note that the states of the above closed loop system ($x$) include the states of the plant ($x_p\in {\rm I\!R}^n$) and possible states of the NN ($x_{nn}\in{\rm I\!R}^b, b\geq 0$)\footnote{These states exist only for RNN (See Sec. \ref{RNNexp}).}. However system requirements are usually on the plant states rather than the NN states, so the value of the NN states $x_{nn}$ do not affect the robustness value directly. As a result $z_*\in {\rm I\!R}^n $ only concerns $x_p$ and any value of $x_{nn}$ is considered to be desired for falsification. In the rest of the paper, the superscript $i$ shows the variables corresponding to the {\it i-th} iteration.

\begin{problem}
\label{pr1}
Given an STL formula $\varphi$, an initial condition $x_p^i(0)$\footnote{The initial condition for neural network states $x_{nn}$ are usually considered as zero.}, and an input signal $w^i$, calculate with respect to the formula $\varphi$, the critical time $t^i_*$ and the critical point $z^i_*$ for the solution to the system of Eq. (\ref{cl}) denoted as $x^i=[x_p^i,x_{nn}^i]$, using the initial condition $x(0)=x^i(0)=[x_p^i(0),zeros(b)]${\footnote{zeros(b) is a vector of b zeros}} and input $w=w^i$. Let $r^i_*\triangleq [z^i_*, x^i_{nn}(t_*^i)]$, and solve the following constrained minimization problem:
\begin{align}\label{optp}
\underset{x_p(0),w}{\text{Minimize}}\quad J^i = \frac{1}{2}&\big( x(t_*^i)-r_*^i\big) ^\top\big(x(t_*^i)-r_*^i\big)\\ \nonumber
s.t &\quad  \dot{x}=f(x,w)\\
x_p&(0)\in X_0, w \in U \nonumber
\end{align}
\end{problem}
%Given a STL formula \phi, assuming that the initial condition x_p^i(0), and the input signal w^i are not the local minimizers of J^i, find \delta x^i(0) and \delta w^i for which there exist a step-size \bar{h}>0 such that for any 0<h<\bar{h}, J^i(x^i)<J^i(x^{i+1}) where x^{i+1} is the solution to (5) with x(0)=x^i(0)+h*\delta x^i(0) and w=w^i+h*\delta w^i.
\subsection{Specification Falsification Attack}
\label{sec}
Due to the nonlinear constraint, finding the global minimizer to Problem (\ref{pr1}) cannot be guaranteed. However, taking a small enough step in the direction of the negative of the gradient of the cost function (\ref{optp}) w.r.t $x_0$ and $w$, will decrease the cost locally. Using the well known method of the Lagrange multipliers, Problem \ref{pr1} can be reduced to the problem of minimizing the following cost function:
%Using the results from well known optimal control methods, we 
\begin{equation*}\vspace{-2pt}
\bar{J}^i = \frac{1}{2}\big( x(t_*^i)-r_*^i\big)^\top\big(x(t_*^i)-r_*^i\big)+\int_{0}^{t_*^i}\lambda ^\top \big(f(x,w)-\frac{dx}{dt}\big)dt 
\end{equation*}
Forming the Hamiltonian as $H(x,w)=\lambda ^\top f(x,w)$, and $\phi^i(x) = \frac{1}{2}\big( x-r_*^i\big)^\top\big(x-r_*^i\big)$, $\bar J^i$ can be written as:
\begin{equation*}
\bar{J}^i = \phi^i(x(t_*^i))+\lambda(0) ^\top x(0)-\lambda(t_*^i) ^\top x(t_*^i) +\int_{0}^{t_*^i} (H(x,w) + \frac{d\lambda}{dt} ^\top x\big)dt 
\end{equation*}
As a result, the gradient of the cost function $\bar J^i$ is: 
\begin{align*}
\delta{\bar J^i} =& \Big(\frac{d\phi^i \big(x^i(t^i_*)\big)}{dx}- \lambda^\top(t_*^i)\Big)\delta x(t_*^i)+\lambda ^\top(0) \delta x(0)\\
&+\int_{0}^{t_*^i} \Big(\big(\frac{\partial H}{\partial x} + \dot \lambda ^\top\big)\delta x +\frac{\partial H}{\partial w} \delta w\Big)dt 
\end{align*}
By updating the co-states $\lambda$ backward in time with the following final value ordinary differential equation,
\begin{align} \label{inja}
&\dot \lambda =  -\frac{\partial H}{\partial x}^\top = -  \frac{\partial f}{\partial x}\Big|_{x^i,w^i}^\top \lambda \\
\lambda(t_*^i) = &\big(\frac{d\phi^i \big(x^i(t^i_*)\big)}{dx}\big)^\top = x^i(t_*^i)-r_*^i 
\end{align}
$\delta{\bar J^i}$ is reduced to $\delta{\bar J^i} =\lambda ^\top(0) \delta x(0)+\int_{0}^{t_*^i} \frac{\partial H}{\partial w} \delta w \;dt$. The following choices of $\delta x(0)$ and $\delta w$ with a small enough positive step size $h$ will result in a negative $\delta{\bar J^i}$ and as a result a decrease in $\bar J^i$:
\begin{align}
 \delta x^i(0) &= -\lambda (0)\\
 \delta w^i(t) = &-\frac{\partial H}{\partial w} = -\frac{\partial f}{\partial w}\Big|_{x^i,w^i}^\top\lambda (t)
 \label{tainja}
\end{align}

In order to find $\delta x(0)$ and $\delta w(t)$ using Eq. (\ref{inja}-\ref{tainja}), we either need to differentiate $f$ w.r.t $x$ and $w$ which requires knowledge about $f$ (or $f_p$ and $\it NN$) or we can use modified version of the successive linearization approach introduced in \cite{yaghoubi2017hybrid}.
Recall that linear approximations of $f$ around operating points can usually be provided. Given $x^i_p(0)$ and $w^i(t)$ assume that we take N time samples on the corresponding trajectory and the following is a linear approximation of Eq. (\ref{cl}) at sample time $t_k\in[0,T]$ ($t_1 = 0, t_N = T$)
\begin{align}
\dot{x} &= A^i_k x+B^i_k w \qquad k ={1,..N}\nonumber 
\end{align}
where $A^i_k, B^i_k$ are constant matrices.
For each time $t\in [t_k,t_{k+1}]$, we calculate the time-varying functions $A^i(t)$ and $B^i(t)$ using $A^i_k$ and $B^i_k$ as follows:
\begin{align}
\alpha_k = \frac{t_{k+1}-t}{t_{k+1}-t_k},& \quad
\alpha_{k+1} =  \frac{t-t_k}{t_{k+1}-t_k}\nonumber \\
A^i(t) = \alpha_k A^i_k +\alpha_{k+1} A^i_{k+1},& \quad B^i(t) = \alpha_k B^i_k +\alpha_{k+1} B^k_{k+1} \label{AB}
\end{align}
and calculate $\delta x(0)$ and $\delta w(t)$ using the following equations
\begin{align}
\lambda(&t_*^i) = x^i(t_*^i)-r_*^i \\
&\dot \lambda =  -  A(t)^\top \lambda \\
 &\delta x^i(0) = -\;\lambda (0)\\
\delta w^i&(t) = - \;B(t)^\top\lambda (t) \label{BA}
\end{align}
The linearization matrices $A_k^i,B_k^i$ can be computed analytically or approximated numerically. Similar to the work in \cite{lioptimal}, using numerical approximaions, our approach can be applied to blackbox systems too. 
The MATLAB \texttt{'Linearize'} command that we use in our implementation can compute the linearizations analytically (using a block-by-block approach) or numerically (using perturbetions) for Simulink models. However it is strongly recommends that the analytical approach is used for Simulink models as it is faster and more accurate.

%Using the above approach means that we don't need white box information about the function $f$ or the structure of the $\it{NN}$. All we need is a tool to extract the linear approximations of $f$ around operating points. Depending on the application the above approach could be advantageous when compared to other  approaches for estimating partial derivatives $\frac{\partial f}{\partial x}$ and $ \frac{\partial f}{\partial w} $. For example the method in \cite{lioptimal} requires $2\times(n+\#$ NN states$)$ simulations in each iteration which can get computationally expensive for high order systems. The advantage of their approach is that it can be applied to black-box models.

Algorithm \ref{alg:GD} describes the process of finding the adversarial inputs and initial conditions. In this algorithm, InBox is a function that saturates its first input argument to lie in the set specified in its second input argument. Note that we can stop the algorithm based on different criteria. The algorithm can be stopped if:
\begin{itemize}
\item A maximum number of iterations is reached.
\item The change in the robustness is less than a minimum value.
\item The change in the initial conditions and inputs are less than a minimum value.
\end{itemize}

\begin{algorithm}[t]
		\caption{ Optimal input and initial condition for falsification }
		\label{alg:GD}
		\begin{algorithmic}[1]
			\Require TL formula $\varphi$, $x^1_p(0)$, $w^1(t)$, $X_0$, $U$, and a tool to extract linearizations of $f$, and initial step size $h_0$, and constant $c>1$.
			\Ensure local optimal initial condition $x_p^*$, local optimal input $w^*$.
			\State Initialize $i = 1, d_* = \infty,h = h_0$
			\State Evaluate the system response $x^i(t)$, and find the corresponding robustness value  $d$, and $t^i_*,r_*^i$.
			\State If $d<d_*$ let $d_* = d$, $x_p^*(0) = x_p^i(0)$,  $w^* = w^i$, and $h = ch $, otherwise let $h =h/c$ and go to step 6.
			\State If $d<0$ ($\varphi$ is falsified): stop and return the corresponding $x_p^*(0), w^*$.
			\State Linearize the system around sample times taken in $[0,t^i_*]$ and evaluate $\delta x^i(0)$ and $\delta w^i$ using equations (\ref{AB}-\ref{BA})
			\State While the stop condition is not active, let $x_p^i(0) = {\mbox InBox}(x_p^i(0)+h\:\delta x_p^i(0),\; X_0)$ \footnotemark and $\forall t\in [0,t_*^i]:\:w^i(t) = {\mbox InBox}(w^i(t)+h\:\delta w^i(t), \;U)$ and go back to step 2.
			\State Let $i = i+1$, $\delta x^i(0) = \delta x^{i-1}(0) $ and $\delta w^{i}=\delta w^{i-1}$

		\end{algorithmic}
    \end{algorithm} \footnotetext{$\delta x_p^i(0)$ is the non NN part of $\delta x^i(0)$}
\subsection{Framework} 
\begin{figure}[t]
\centering
\includegraphics[width=.7\linewidth]{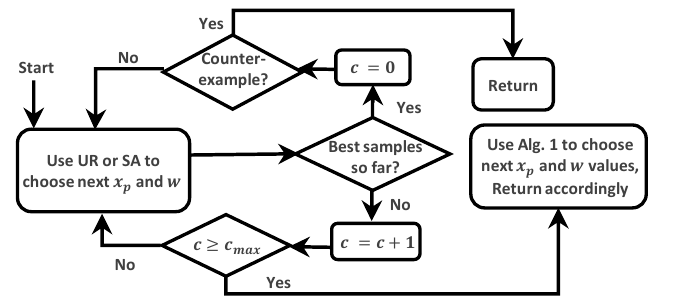}
\caption{The falsification framework}
\label{frame}
\end{figure}

The robustness function is a non-convex non-differentable function in nature. In order to locally solve the problem we defined the function $J_{x_p(0),w}$. However, in order to search for the global minimizer of the robustness function, the gradient based local search still needs to be combined with a ``sampling method for coverage" or a ``stochastic global optimization" approach.
In what follows we combine the local search with Uniform Random Sampling (UR) and Simulated Annealing optimization (SA). The framework is shown in Fig. \ref{frame} where $c=0$ at start and $c_{max}$ is a design choice.

\section{Case Studies}
In this section we study two systems containing NNs. The NNs serve as controllers and they are trained to replicate the behavior of well-known controllers. Motivated by the fact that Simulink models are widely used in industry for modeling complicated systems, both of our case studies are Simulink models that are treated as gray-box, and the only information that we extract from their model, is the dynamical model linearizations along systems' trajectories that are anyway extractable using the Simulink's linear analysis toolbox.
\subsection{Nonlinear system with FNN controller}
\begin{figure}[t]
\centering
\begin{subfigure}
\centering
\includegraphics[width=.67\linewidth]{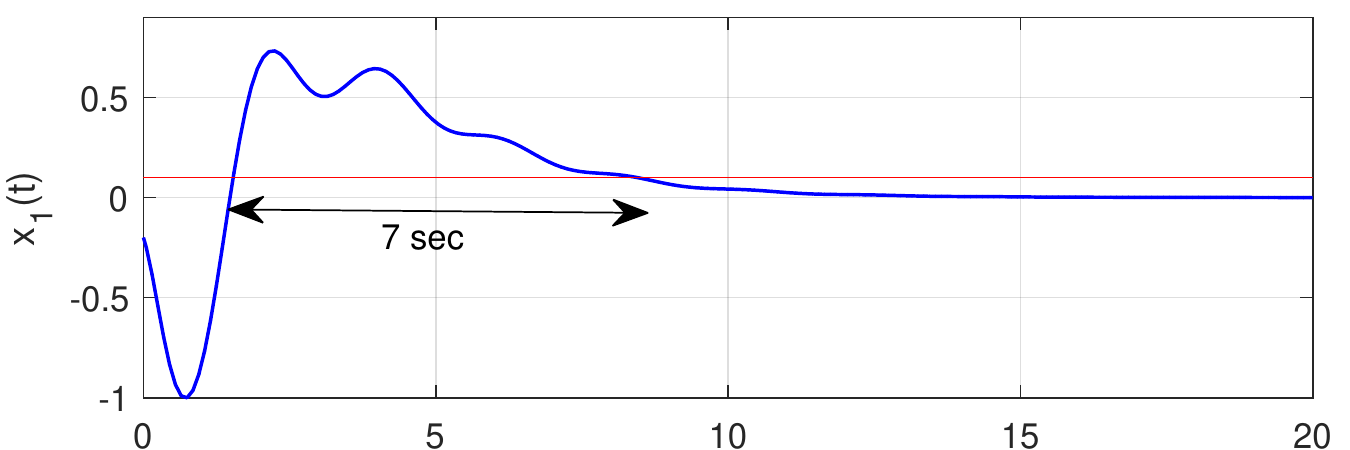}
\end{subfigure}\vspace{-5pt}
\begin{subfigure}
\centering
\includegraphics[width=.67\linewidth]{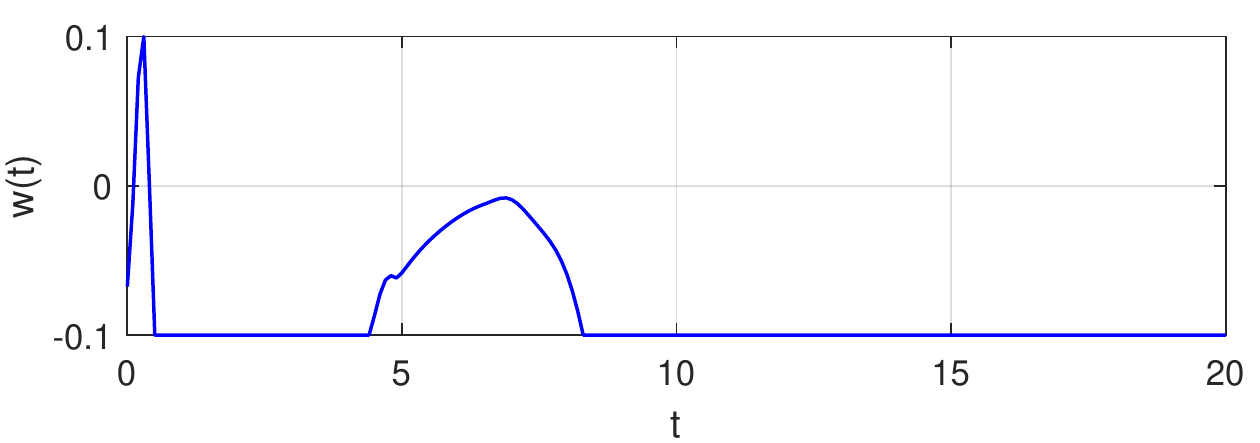}
\vspace{-10pt}
\end{subfigure} 
\caption{Falsifying Input and trajectory: The trajectory does not settle down below 0.1 within 7 seconds of the rise time.}
\label{FNN}
\end{figure}

Consider the following nonlinear system under a FNN controller that has 5 layers and  tangent-sigmoid activation functions. Also let $x_1(0) = -0.2 ,x_2(0) = 5$, and $w(t)\in[-0.1,0.1]$:
\begin{align*}
&\dot x_1 = -0.5 x_1-2e^{-0.5t}sin(3t)+sin(x_2)\\
\dot x_2& = -x_2 + x_1^2(cos(x_2+w(t))+\it{FNN}(x_1,x_2)
\end{align*}
The system is tested against the specification: 
\begin{equation*}
    \Box \big((x_1(t)<0 \wedge \Diamond_{[0,\epsilon]}\; x_1(t)>0) \rightarrow \Diamond_{[0,7]} \Box (x_1(t)<0.1)\big)
\end{equation*}

in which $\epsilon$ is a small positive constant. The requirement requires the signal to always stay below 0.1 within 7 second of the rise time. Starting from $w(t) = 0$ the local optimal search finds an input (shown in Fig. \ref{FNN}) that falsifies the requirement. The robustness for the falsifying trajectory is $-7.7\times 10^{-7}$.

\subsection{Steam Condenser with RNN Controller}\label{STRNN}
\begin{figure}[t]
\centering
\includegraphics[width=.65\linewidth]{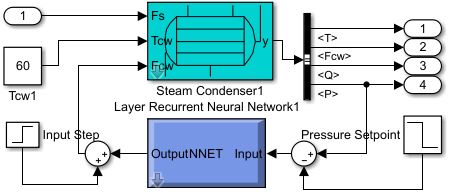}
\caption{Simulink model for Steam Condenser with Feedback RNN Controller}
\label{fig:2}
\end{figure}

\begin{figure}[t]
\centering
\begin{subfigure}
\centering
\includegraphics[width=.68\linewidth]{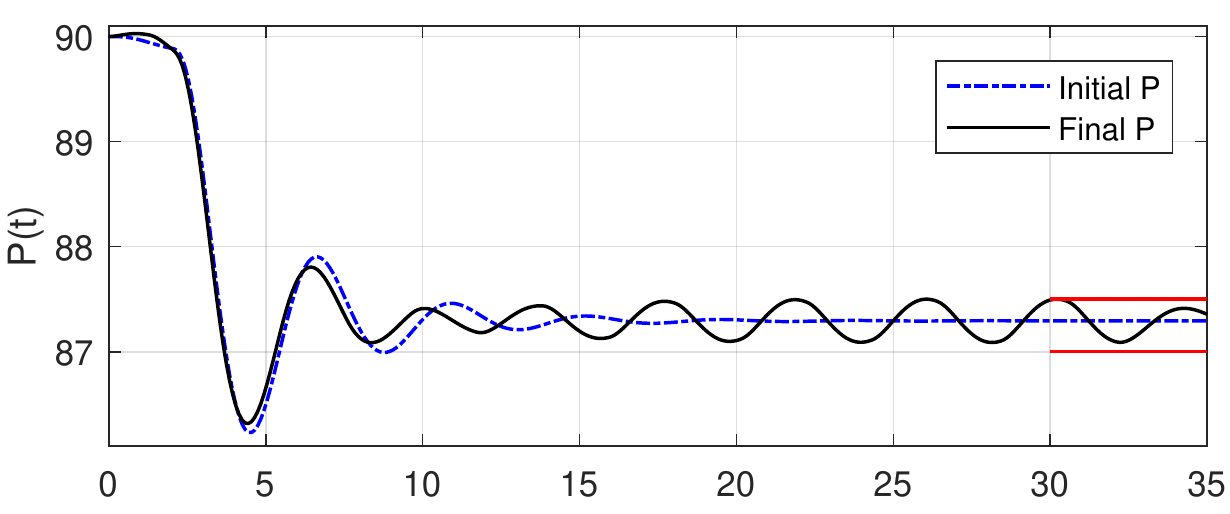}
\end{subfigure}\vspace{-2pt}
\begin{subfigure}
\centering
\includegraphics[width=.7\linewidth]{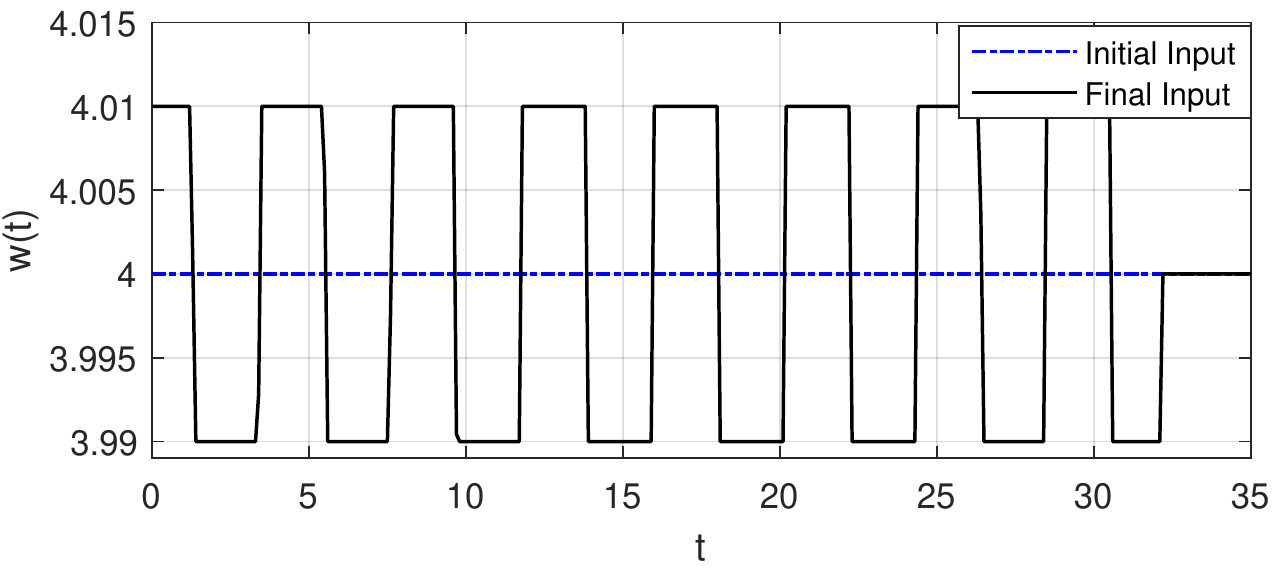}
\end{subfigure} \vspace{-15pt}
\caption{The system robustness is reduced from 0.20633 using a constant input $w(t)=4$ to 0.00030222 using the local optimal input shown in the picture.}
\label{pu}
\end{figure}

We studied a dynamic model of a steam condenser with 5 continuous states based on energy
balance and cooling water mass balance \cite{steam} under an RNN controller with 6 discrete states and tangent-sigmoid activation functions. The Simulink model for the system is shown in Fig. \ref{fig:2}. The steam flow rate $w(t)$ (Input 1 in Fig. \ref{fig:2}) is allowed to vary in the set $[3.99,4.01]$ and the system is tested for $T=35$ seconds against the specification $    \Box_{[30,35]} \:p(t)\in [87,87.5]$.
Starting from a constant valued signal $w(t)= 4$ that results in a robustness value equal to 0.20633, the above approach finds a falsifying trajectory with robustness 0.00030222. The initial and final trajectories and inputs are shown in Fig. (\ref{pu}).
Using $w(t)= 3.99$ and  $w(t)= 4.01$ initially, the robustness values were reduced from 0.24131 to 0.00033674 and from 0.17133 to 0.0002290, respectively. While the local search reduces the robustness values significantly in all the above 3 cases, in none of them a falsifying behavior is found. The importance of combining this local search with a global sampler/optimizer becomes more clear in the next section where the combination of the local search with uniform random sampling or Simulated Annealing method finds adversarial examples.

Note that, while the utilized NNs have fairly small number of layers (since they were found to perform good enough during the training phase), the scalability of the proposed approach was tested on the systems of sections 5.1 and 5.2 including NN controllers with much larger number of layers.  These experiments showed that the proposed approach scales well. Theoretically increasing the number of layers/neurons in FNNs or the number of non-recurrent layers (with no delay/memory) in RNNs will just increase the number of blocks in the Simulink model linearly. Since MATLAB analytical linearization is computed block-by-block, increasing the number of these kinds of layers increase the linearization complexity by $O(n)$. 
However increasing the size of state-space or the number of layers of the RNN that include delay increases the linearziation complexity faster. Specifically the size of linearized matrices grow with $O(n^2)$ by the number of state-space or RNN states.  
However in practice, we observed much less increase in the computation time of the overall algorithm when increasing the size of the NNs and system states.

\section{EXPERIMENTAL RESULTS}
Experiments are conducted using MATLAB 2017a on an Intel(R) Core(TM) i7-4790 CPU @3.6 GHZ with 16 GB memory processor with Windows 10 Enterprise. 
We used Uniform Random Sampling (UR) and Simulated Annealing (SA) implementations of S-Taliro \cite{annpureddy2011s} unaided and aided by the optimal local search (UR+GD and SA+GD, respectively) for finding adversarial inputs to the problem described in Sec. \ref{STRNN} with RNN in the loop.
In the UR+GD implementation, local optimal search is performed when the sampler cannot find a sample with a less robustness value 50 times in a row, and in the SA+GD implementation it is applied when the optimizer cannot find a less robust sample 30 times in a row. We run the experiments 50 times, and in each run the maximum execution time is limited to 60 seconds\footnote{The next sample is taken only if the execution time so far is less than 60 seconds. The algorithm returns faster in case of finding a counter/adversarial example.}. The search is initialized with the same seed for all the experiments. The above search methods are compared against the number of falsifications found, average minimum robustness found, average execution time, and average total number of simulations before returning. The improvement in the results from left to right in Table \ref{tab:commands} is evident and it motivates the use of the proposed local search. While SA and UR were not able to find any counterexamples in 50 runs, their combination with gradient based descent found an adversarial example in all the runs within a short amount of time and with less than 90 simulations on average.
\begin{table}
  \caption{Falsification Results of Steam Condenser system with RNN controller using different search methods.}\vspace{-5pt}
  \label{tab:commands}
  \begin{tabular}{ccccl}
    \toprule
       & UR & SA & UR+GD & SA+GD\\
    \midrule
    \# falsifications & 0/50 & 0/50 & 50/50 & 50/50\\
    Avg min robustness & 0.0843&  0.0503 & -0.0018 &  -0.0016\\
    Avg execution time & >60 & >60  & 15.7812 & 13.0688\\
    Avg \# simulations & 600& 600  & 87.48 &62.26\\
    \bottomrule
  \end{tabular}
\end{table}

\section{CONCLUSIONS and future work}

In this paper, we proposed a gradient based local search approach for falsification using results from optimal control theory. 
We applied our method to a case study of a Simulink model of a steam condenser with an RNN controller. 
The results suggest that when combined with a sampler or a global optimizer, the gradient based local search is indeed very useful in detecting counterexamples to the requirements for control systems with RNNs.

As a future line of work, we will extend our results to falsification of dynamical systems with classifier neural networks. This can be done by manipulating the system's inputs in a way that the NN inputs approach the decision boundary of the falsifying NN output. 
% The idea is illustrated in Fig. \ref{fig:3}. 
In a different line of work we will try to augment the adversarial samples into the training set of the NN in order to adapt its gains and improve the system's performance.

\section*{Acknowledgement}{This work was partially supported by the NSF awards CNS 1350420, IIP-1361926, and the NSF I/UCRC Center for Embedded Systems.}
\bibliographystyle{ACM-Reference-Format}
\bibliography{sample-bibliography}

\end{document}